%% file: root.tex
\documentclass[letterpaper, 10 pt, conference]{ieeeconf}  % Comment this line out if you need a4paper

\IEEEoverridecommandlockouts                              % This command is only needed if 
\usepackage{amsmath,amssymb,amsfonts}
\usepackage{bm}
\usepackage{url}
\usepackage{graphics}
\usepackage{algorithmic}
\usepackage{multirow}
\usepackage{booktabs}
\usepackage{multirow}
\usepackage[ruled,linesnumbered]{algorithm2e}
\usepackage{textcomp}
\usepackage{tabularx}
\usepackage{relsize}
\usepackage[flushleft]{threeparttable}
\usepackage{cite}
\usepackage[table,dvipsnames]{xcolor}
\usepackage{colortbl}
\usepackage{enumerate}
\let \labelindent \relax 
\usepackage[inline]{enumitem}
\usepackage{acronym}
\usepackage{siunitx}
\usepackage{subfig}
\usepackage{comment}
\usepackage{caption}
\usepackage{wrapfig}
\usepackage{graphicx}
\usepackage{subfig}

\input{macros}

\makeatletter
\let\NAT@parse\undefined
\renewcommand{\thefigure}{\arabic{section}.\arabic{figure}}
\makeatother
\usepackage{cite} 
\usepackage{hyperref} 

\makeatletter
\renewcommand{\thefigure}{\arabic{figure}}
\makeatother

\title{\LARGE \bf
GaussTwin: Unified Simulation and Correction with Gaussian Splatting for Robotic Digital Twins}

\author{
    Yichen Cai$^{1}$, Paul Jansonnie$^{15}$, Cristiana de Farias$^{1}$,  Oleg Arenz$^{1}$, Jan Peters$^{1234}$% <-this % stops a space
    \thanks{
    This work was funded by the German Research Foundation (DFG) - Project number PE 2315/18-1,  the EU’s Horizon Europe project ARISE - Grant number 101135959, and partially supported by the German Federal Ministry of Research, Technology and Space (BMFTR) under the Robotics Institute Germany (RIG).
    This work was also supported by a hardware donation from NVIDIA through the Academic Grant Program, and by the Lichtenberg high-performance computer of TU Darmstadt. 
    \newline
    \hspace*{1em} We acknowledge that ChatGPT is used for grammar enhancement, and Copilot is used to generate some auxiliary functions in the code.
    \newline
    \hspace*{1em}
    $^{1}$Intelligent Autonomous Systems Lab, Technical University of Darmstadt, Germany.
    $^{2}$Hessian.AI, Germany.
    $^{3}$German Research Center for AI (DFKI), SAIROL, Germany.
    $^{4}$Robotics Institute Germany (RIG).
    $^{5}$NAVER LABS Europe.
    \newline \hspace*{1em} Corresponding author: {\tt\small yichen.cai@tu-darmstadt.de }
    }
}

\begin{document}\sloppy

%% ---------------------------------- Start ------------------------------ %%

\maketitle
\thispagestyle{empty}
\pagestyle{empty}
%%%%%%%%%%%%%%%%%%%%%%%%%%%%%%%%%%%%%%%%%%%%%%%%%%%%%%%%%%%%%%%%%%%%%%%%%%%%%%%%
\input{acronyms}
\input{sections/0_abstract}
%%%%%%%%%%%%%%%%%%%%%%%%%%%%%%%%%%%%%%%%%%%%%%%%%%%%%%%%%%%%%%%%%%%%%%%%%%%%%%%%

\input{sections/1_introduction}
\input{sections/2_related_works}
\input{sections/3_preliminary}

\input{sections/4_methodology}
\input{sections/5_experiments}
\input{sections/6_conclusion}

\bibliographystyle{IEEEtran}
\bibliography{library}

%%%%%%%%%%%%%%%%%%%%%%%%%%%%%%%%%%%%%%%%%%%%%%%%%%%%%%%%%%%%%%%%%%%%%%%%%%%%%%%%
%%% Uncomment if needed
% \section*{APPENDIX}

% Appendixes should appear before the acknowledgment.

% \section*{ACKNOWLEDGMENT}

% The preferred spelling of the word ÒacknowledgmentÓ in America is without an ÒeÓ after the ÒgÓ. Avoid the stilted expression, ÒOne of us (R. B. G.) thanks . . .Ó  Instead, try ÒR. B. G. thanksÓ. Put sponsor acknowledgments in the unnumbered footnote on the first page.

%%%%%%%%%%%%%%%%%%%%%%%%%%%%%%%%%%%%%%%%%%%%%%%%%%%%%%%%%%%%%%%%%%%%%%%%%%%%%%%%

% \begin{thebibliography}{99}
% \bibitem{c1} G. O. Young, ÒSynthetic structure of industrial plastics (Book style with paper title and editor),Ó 	in Plastics, 2nd ed. vol. 3, J. Peters, Ed.  New York: McGraw-Hill, 1964, pp. 15Ð64.
% \end{thebibliography}

\end{document}

%% file: macros.tex
\newcommand{\mug}{\texttt{mug}}
\newcommand{\spamcan}{\texttt{spam can}}
\newcommand{\tblock}{\texttt{T-shaped block}}
\newcommand{\rubik}{\texttt{Rubik’s Cube}}
\newcommand{\rope}{\texttt{rope}}

\definecolor{softred}{RGB}{220,100,80}
\definecolor{softpurple}{RGB}{150,80,190}
\definecolor{softmagenta}{RGB}{190,70,170}
\definecolor{magenta}{RGB}{220,40,190}

\newcommand{\reshighlight}[1]{\textbf{\textcolor{magenta}{#1}}}

%% file: acronyms.tex
\newacro{3DGS}[3DGS]{3D Gaussan Splatting}
\newacro{PBD}[PBD]{Position-Based Dynamics}
\newacro{DLO}[DLOs]{Deformable Linear Objects}
\newacro{GT}[GaussTwin]{Gaussian Splatting for Robotic Digital Twins}
\newacro{PEGS}[PEGS]{Physically Embodied Gaussian Splatting} 
\newacro{RBD}[RBD]{Rigid Body Dynamics } 

%% file: sections/0_abstract.tex
\begin{abstract}
Digital twins promise to enhance robotic manipulation by maintaining a consistent link between real-world perception and simulation. However, most existing systems struggle with the lack of a unified model, complex dynamic interactions, and the real-to-sim gap, which limits downstream applications such as model predictive control. Thus, we propose GaussTwin, a real-time digital twin that combines position-based dynamics with discrete Cosserat rod formulations for physically grounded simulation, and Gaussian splatting for efficient rendering and visual correction. By anchoring Gaussians to physical primitives and enforcing coherent SE(3) updates driven by photometric error and segmentation masks, GaussTwin achieves stable prediction–correction while preserving physical fidelity. Through experiments in both simulation and on a Franka Research 3 platform, we show that GaussTwin consistently improves tracking accuracy and robustness compared to shape-matching and rigid-only baselines, while also enabling downstream tasks such as push-based planning. These results highlight GaussTwin as a step toward unified, physically meaningful digital twins that can support closed-loop robotic interaction and learning. 
Code and videos are available at \url{https://6cyc6.github.io/gstwin/}.

\end{abstract}

%% file: sections/1_introduction.tex
\section{Introduction}
\label{sec:intro}

Building a real-time, dynamic digital twin offers significant benefits for robotic manipulation.
Unlike traditional simulation, which creates a virtual environment separate from the real world, a dynamic digital twin predicts future states, continuously corrects and synchronizes the robot with reality, and generates visual representations of changing environment conditions.
This capability helps bridge the real-to-simulation gap, enabling more effective planning, tracking, and the generation of robot–object interaction videos for policy learning \cite{yu2025real2render2real, abou2025real}.
However, developing digital twins that rely on a unified physics model capable of handling a wide range of properties and behaviors, such as rigid and deformable bodies, contact interactions, and different material types, remains a significant challenge.

In practice, digital twins have typically been constructed using point clouds \cite{schulman2013tracking, tang2017state, liu2021real, liang2024real}, meshes \cite{petit2015real}, or NeRF \cite{abou2024particlenerf} to represent the 3D environment, but these representations face trade-offs among efficiency, differentiability, and fidelity. Recently, \acf{3DGS}~\cite{kerbl20233d} has emerged as an alternative for 3D scene representation. By modeling a scene as a dense set of 3D Gaussians,  \ac{3DGS} enables efficient and differentiable novel-view image synthesis with low memory cost and high visual fidelity. Building on this idea, works such as,~\cite{zhang2024dynamics, jiang2025phystwin, zhang2025particle} combine \ac{3DGS}-based reconstruction with neural dynamic models to construct more visually accurate digital twins that capture dynamics directly from videos. Here, by integrating Graph Neural Networks with a particle-grid representation,~\cite{zhang2025particle} demonstrated strong capabilities in modeling deformable object dynamics, achieving accurate long-horizon predictions while leveraging \ac{3DGS} for rendering. However, these learning-based approaches often struggle to generalize beyond their training distribution and demand extensive, labor-intensive data collection and processing.
\begin{figure}
    \centering
    \includegraphics[width=1\linewidth]{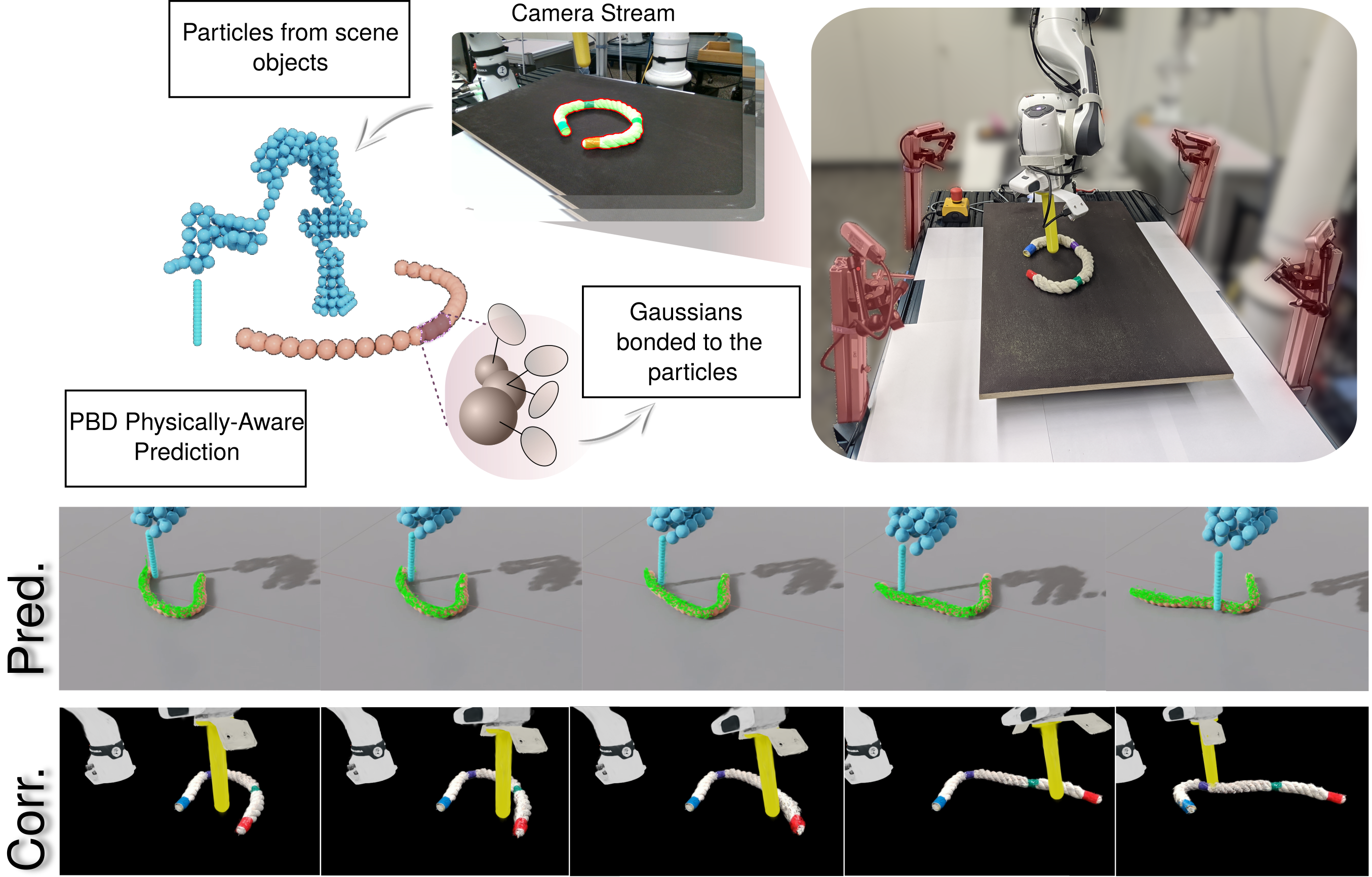}
\caption{Tracking performance of GaussTwin on rigid objects and DLO. The system takes multi-view camera observations as input. First, the objects in the scene are masked, represented by particles, and bonded to 3D Gaussians. The motion of these particles is then predicted at each time step using PBD, and subsequently refined through Gaussian splatting optimization. The bottom two rows illustrate the prediction–correction process carried out by GaussTwin.}
    \label{fig:pipeline}
\end{figure}

Concurrently with learning-based methods, another line of work has explored hybrid approaches that integrate physics-based simulation with visual or data-driven corrections. As was often the case in earlier digital twin implementations, point clouds or signed distance fields were commonly used for the visual correction step~\cite{schulman2013tracking, tang2017state, liu2021real, xiang2023trackdlo, liang2024real}. However, these representations are typically tailored to specific object types such as ropes or soft tissues, which restricts their general applicability. In addition, both point clouds and SDFs are prone to occlusion and sensor noise. In parallel with the advances described above, \ac{3DGS} has also emerged as a promising alternative for the visual correction component of prediction-correction models. For instance, \cite{abou2024physically, abou2025real} introduced a \ac{3DGS}-based visual corrective world model, which uses a simulator for prediction and then refines the results through visual feedback. Both methods are built upon \acf{PBD}~\cite{muller2007position}, a simulation framework that enforces constraints on positions of particles to achieve fast and stable real-time dynamics. Specifically, \cite{abou2024physically} proposed a hybrid Particle-Gaussians representation, where particles evolve in a PBD simulator while bonded 3D Gaussians are used for image rendering and visual feedback. Their approach further employed a shape-matching algorithm with oriented particles~\cite{muller2005meshless, muller2011solid} to simulate both rigid and deformable objects within the PBD framework. However, the shape-matching algorithm lacks physically meaningful properties, leading to inaccurate simulation predictions, especially for \acf{DLO}, e.g., a rope. 
Moreover, it requires a high visual correction gain because the Gaussians are optimized independently, which often leads to strong oscillations. 
In follow-up work,~\cite{abou2025real} adopted rigid-body dynamics to achieve more accurate predictions and applied their improved corrective model to a manipulation task, where they used it to push a T-shaped object. While this enhanced rigid-body tracking, it came at the cost of losing the ability to simulate and track deformable objects. 

In this paper, we present \acf{GT}, a \emph{unified hybrid framework} that overcomes the limitations of prior \ac{3DGS}-based corrective models (See Fig.~\ref{fig:pipeline}). 
Instead of relying on shape matching alone or restricting to rigid-body dynamics, we extend \ac{PBD} with the discrete Cosserat rod model~\cite{kugelstadt2016position}, which provides a physically meaningful formulation capable of accurately modeling both rigid objects and \ac{DLO} within the same framework. 
To further improve stability, we introduce a joint optimization scheme for 3D Gaussians that constrains them to move coherently with their associated rigid bodies or rod segments, preventing independent drift that previously caused oscillations and required high correction gains. 
Together, these contributions enable GaussTwin to unify the prediction and correction of rigid and deformable objects, while ensuring more stable, efficient, and physically accurate digital twin simulations. 
The key contributions of this paper are summarized as follows:

\begin{itemize}

\item We introduce \ac{GT}, a hybrid framework that combines \ac{PBD} with 3D Gaussians to jointly predict and correct the state of both rigid bodies and \ac{DLO}, thereby bridging the real-to-sim gap.
\item By leveraging segmentation masks, enforcing coherent motion of 3D Gaussians with their corresponding rigid bodies or rod segments, and incorporating physically plausible constraints in \ac{PBD}, our method achieves stable and precise prediction–correction without sacrificing real-time performance. 
\item Through simulation and real-world experiments, we demonstrate that \ac{GT} achieves more accurate and robust tracking compared to prior \ac{3DGS}-based corrective models~\cite{abou2024physically}.
\item We show the effectiveness of our method in a downstream robotic planning task, highlighting its potential to support closed-loop interaction and control in real-world environments.

\end{itemize} 

%% file: sections/2_related_works.tex
\section{Related Works}
\label{sec:related_works}
Building a model that can simulate and track both rigid and linear deformable objects is crucial for complex robotic manipulation tasks.
Many physics simulators are available to simulate rigid bodies \cite{todorov2012mujoco, NVIDIA_Isaac_Sim} and DLOs \cite{NVIDIA_Isaac_Sim, laezza2021reform, yang2022learning, warp2022}.
However, they fail to predict and track states of the objects over a long horizon due to mismatches with the real world, especially for objects with unknown physical parameters.
To address this, some works estimate physical parameters through reconstruction methods, thereby helping bridge the sim-to-real gap.
Most early works estimate parameters for objects with known geometry using synthetic data \cite{geilinger2020add, du2021diffpd, qiao2021differentiable} or point clouds \cite{sundaresan2022diffcloud, liu2023robotic}.
With the development of 3D reconstruction techniques \cite{mildenhall2021nerf, kerbl20233d} in computer vision, recent works leverage NeRF or Gaussian Splattings to reconstruct the geometry and texture of objects and to estimate parameters from videos.  
However, these methods remain challenging for estimating the parameters of objects with complex physical properties, such as DLOs.
In addition, they are limited to specific motions and scenarios.
Another branch of work \cite{evans2022context, yan2020self, ai2024robopack, shi2024robocraft, zhang2024dynamics, jiang2025phystwin, zhang2025particle} employs data-driven methods, which train a deep neural network to learn dynamics directly from videos.
By incorporating physical priors, \cite{jiang2025phystwin, zhang2025particle} propose a physics-informed graph neural network that learns deformable object dynamics from sparse real-world robot interaction videos.
When combined with 3DGS, the model achieves improved reconstruction and prediction performance across a range of objects and motions.
However, these methods still require building a complex system to collect and preprocess training data.
Additionally, they struggle to generalize to out-of-training scenarios involving external disturbances, collisions, and interactions among multiple objects.

Recently, \cite{abou2024physically, abou2025real} proposed corrective world models.
By integrating real-world visual feedback into the simulator, these methods construct a real-time digital twin that effectively bridges the real-to-sim gap.
Furthermore, \cite{liu2021real, liang2024real} build a PBD simulation to simulate soft tissues and leverage the point cloud observation to correct the predicted state of the simulation.
Although they achieved improved tracking performance, perception noise and computational cost limited their usage.
Most related to our method, \cite{abou2024physically} builds a hybrid particle-Gaussians model from RGB-D images and initializes it in a PBD simulation. 
In this case, future states of the particle-Gaussians model are initially predicted within the PBD simulator.
Visual forces derived from rendered images of the bonded 3D Gaussians are applied to the particles to correct their states.
However, they use shape matching to ensure that unconnected particles retain the shape of the object they represent, which has no physical properties of its own.
Thus, the simulation's prediction might introduce large errors, especially for DLOs.
In \cite{abou2025real}, they switch to rigid body dynamics, but lose the ability to track DLOs.
\cite{dinkel2025dlo} uses a similar method to track DLOs, but the simplified rope model restricted their tracking performance. 
We extend their method and build a unified PBD simulation capable of modeling and tracking both rigid objects and DLOs.

%% file: sections/3_preliminary.tex
\section{Preliminary}
\label{sec:priliminary}
\subsection{Position-Based Dynamics Simulation}
\label{sec:priliminary_PBD}
\acf{PBD} \cite{muller2007position} has been extensively used for building interactive physics systems due to its simplicity, stability, visual plausibility, and computational speed \cite{bender2017survey}. 
These properties make it especially suitable for building a real-time visual corrective digital twin.
In general, \ac{PBD} is a particle-based system paired with various constraints based on the dynamics of the objects. 
By integrating rigid body dynamics \cite{deul2016position} and the discrete Cosserat rod model \cite{kugelstadt2016position}, \ac{PBD} can be extended to a unified framework that jointly simulates different types of objects, including rigid bodies and \ac{DLO}.
The overall \ac{PBD} simulation process is outlined in Algorithm~\ref{alg:PBD_flowchart}. It begins with system initialization, where the state of each rigid body is defined. Specifically, the $i$\textit{-th} rigid body is represented by its mass $m_{i}$, position $\bm{x}_{i}$, velocity $\bm{v}_{i}$, inertial matrix $\bm{I}_{i}$, orientation $\bm{q}_{i} \in \bm{SO}(3)$, and angular velocity $\bm{\omega}_{i}$.
Following the Cosserat rod model, a DLO (such as a rope) can be approximated by a sequence of linear segments along its centerline. The state of a DLO is therefore defined by two sets of variables: positional variables, including the mass $m_{i}$, position $\bm{x_{i}}$, and velocity $\bm{v_{i}}$ of each particle along the centerline; and rotational variables, including the inertia matrix $\bm{I}_{i}$, orientation $\bm{q}_{i}$, and angular velocity $\bm{\omega}_{i}$ of each segment. 
\begin{algorithm}[t]
\small
\caption{Unified PBD Simulation}
\label{alg:PBD_flowchart}
\begin{algorithmic}[1]
\STATE InitializeRigidBodyStates()
\STATE InitializeRopeStates()
% \STATE $h \gets \Delta t / \mathrm{numSubsteps}$
\WHILE{simulation not stopped}
  \FOR{$i = 1$ \TO $\mathrm{numSubsteps}$}
    \STATE CollectCollisionPairs()
    \STATE StatePrediction()
    \FOR{$j = 1$ \TO $\mathrm{solverIterations}$}
      \STATE ClearDeltas()
  \STATE $(\Delta\bm{x}^{C}, \Delta\bm{q}^{C}) \gets$ SolveRigidParticleContact() ~\eqref{eqa:rp_solve}
      \STATE $(\Delta\bm{x}^{S}, \Delta\bm{q}^{S}) \gets$ SolveShearStretch() ~\eqref{eqa:solve_shear}
      \STATE $(\Delta\bm{x}^{B}, \Delta\bm{q}^{B}) \gets$ SolveBendTwist() ~\eqref{eqa:solve_bend}
      \STATE $(\Delta\bm{x}^{R}, \Delta\bm{q}^{R}) \gets$ SolveRigidBodyContact() ~\eqref{eqa:rb_solve}
      \STATE ApplyDeltas()
    \ENDFOR
    \STATE UpdateVelocities()
  \ENDFOR
\ENDWHILE
\end{algorithmic}
\end{algorithm}
After initialization, the simulation enters its main loop, which is executed over several substeps.
Each substep begins with a collision check to build the contact constraints, after which PBD predicts the positions and orientations using semi-implicit Euler integration.
Next, a Jacobian solver is used to solve the constraints $\bm{C}(\bm{p}_0,...,\bm{p}_n)$, which is the core step of the algorithm. Here, each
$\bm{p}_i$ denotes either a position or an orientation.
The constraints may be either an equality or an inequality, scalar or vector.
To compute the update displacements, each constraint is locally linearized as $\bm{C}(\bm{p} + \Delta \bm{p}) 
= \bm{C}(\bm{p}) + \nabla_{\bm{p}} \bm{C} \, \Delta \bm{p} = 0$.
The displacement direction is restricted in the projected space and can be solved using a Lagrangian multiplier
\begin{equation}
\begin{aligned}
\label{eqa:PBD_update}
\bm{\lambda}
&= - \left(
    \sum_k 
        \left( \nabla_{\bm{p}_k} \bm{C} \right)
        \bm{W}_k
        \left( \nabla_{\bm{p}_k} \bm{C} \right)^{\!T}
\right)^{\!-1}
\bm{C}(\bm{p}), \\ 
\Delta \bm{p}_i
&= \bm{W}_i 
  \left( \nabla_{\bm{p}_i} \bm{C} \right)^{\!T}
  \bm{\lambda},
\end{aligned}
\end{equation}
where $k$ is the number of states involved in a constraint, and $\bm{W}=\bm{diag}[m_{1}^{-1},...m_{n_x}^{-1},\bm{I}_{1}^{-1},...\bm{I}_{n_q}^{-1}]$ encodes the inverse masses and inertias. We note that, each variable update is weighted by $\bm{W}_i$ to ensure conservation of momentum. Since all constraints are independent, they can be efficiently solved in parallel on a GPU.
The accumulated updates are then applied to the predicted state and cleared before the next solver iteration.
Finally, the corrected predictions are used to update the velocities.
Further details on the constraint formulations can be found in Section~\ref{sec:PBD_constraints}, and a comprehensive discussion of PBD is provided in \cite{muller2007position}.

\subsection{3D Gaussian Splattings}
3D Gaussian Splatting \cite{kerbl20233d} explicitly represents the scene by a set of 3D Gaussians.
Each Gaussian $\bm{g}_i$ is defined by its mean $\bm{\mu}_{i}\in\mathbb{R}^{3}$, covariance matrix $\bm{\Sigma}_{i}\in\mathbb{R}^{3\times 3}$, opacity $\alpha_{i}\in\mathbb{R}$, and color $\bm{c}_i\in\mathbb{R}^{3}$.
The covariance matrix is decomposed into a rotation matrix $\bm{R}_{i}$ and a diagonal scale matrix $\bm{S}_{i}$ by $\bm{\Sigma}_{i}=\bm{R}_{i}\bm{S}_{i}\bm{S}_{i}^{T}\bm{R}_{i}^{T}$. 
For the training, the rotation is reparameterized as a unit quaternion $\bm{q}_i$, and the diagonal elements of the scale matrix are formed as a scale vector $\bm{s}_{i}$.
Given a camera configuration, the Gaussians are projected onto the image plane and sorted by depth. 
The image is then rendered by $\alpha$-blending.

%% file: sections/4_methodology.tex
\section{Methodology}
In this section, we present the main aspects of \ac{GT}, which allows us to align simulation with visual correction to create a physically accurate digital twin for manipulation tasks. 
An overview of the pipeline is shown Fig.~\ref{fig:pipeline}.
An object-Gaussian bond is initially constructed from RGB-D images captured by multiple cameras. 
After initialization, the tracking, which consists of a prediction step and a correction step, runs at 25 Hz.
We first describe the constraints used in the PBD simulation and then discuss the details of the initialization and the tracking procedures. 

\subsection{Constraint definition}
\label{sec:PBD_constraints}
To simulate rigid bodies in \ac{PBD}, we employ a number of different constraints. For contact collision constraints between the rigid bodies, we have
\begin{equation}
\label{eqa:rb_constraint}
C^{R}(\bm{x}_i, \bm{q}_i, \bm{x}_j, \bm{q}_j;\bm{n},\bm{b}_i,\bm{b}_j) = \bm{n}\cdot(\bm{b}_j-\bm{b}_i) \geq 0,
\end{equation}
where $\bm{n}$ is the contact normal, $\bm{b}_i$ is the contact position of the rigid body $i$ in the world frame, and the indices $i$ and $j$ denote the two rigid bodies involved in the contact.
Here, the generalized inverse masses are
\begin{equation}
\begin{aligned}
w_i &= m_i^{-1} + (\bm{r}_i \times \bm{n})^\top \bm{I}_i^{-1} (\bm{r}_i \times \bm{n}), \\
w_j &= m_j^{-1} + (\bm{r}_j \times \bm{n})^\top \bm{I}_j^{-1} (\bm{r}_j \times \bm{n}), 
\end{aligned}
\end{equation}
where $\bm{r}_i$ and $\bm{r}_j$ are vectors from the center of mass to the contact position.
The resulting state correction of each rigid body is then given by
\begin{equation}
\begin{aligned}
\Delta\bm{x}_{i}^{R} &=-\frac{m_i^{-1}}{w_i + w_j}C^{R}\bm{n}, 
\:\Delta\bm{x}_{j}^{R} =\frac{m_j^{-1}}{w_i + w_j}C^{R}\bm{n}, \\
\Delta\bm{q}_{i}^{R} &=-\frac{1}{2}[\bm{I}_i^{-1}(\bm{r}_i\times 
\frac{1}{w_i + w_j}C^{R}\bm{n}),0]\bm{q_{i}},\\
\Delta\bm{q}_{j}^{R} &=\frac{1}{2}[\bm{I}_j^{-1}(\bm{r}_j\times 
\frac{1}{w_i + w_j}C^{R}\bm{n}),0]\bm{q_{j}}.
\label{eqa:rb_solve}
\end{aligned}
\end{equation}

The next constraint is the collision between a rigid body $i$ and a particle $j$, formulated as $C^{C}(\bm{x}_{rb_i}, \bm{q}_{rb_i}, \bm{x}_j;\bm{n},\bm{b}, r_j) = \bm{n} \cdot (\bm{x}_{rb_i} - \bm{b}) - r_j \geq 0 $. This constraint is solved in a manner similar to the contact collision constraints between rigid bodies. 
The only difference is that $\omega_j$ is simplified to $m_j^{-1}$ and there is no orientation update for the particle.
In this case, the update rule is as follows:
\begin{equation}
\begin{aligned}
\Delta\bm{x}_{i}^{C} &=-\frac{m_i^{-1}}{w_i + w_j}C^{C}\bm{n}, \:\Delta\bm{x}_{j}^{C} =\frac{m_j^{-1}}{w_i + w_j}C^{C}\bm{n}, \\
\Delta\bm{q}_{i}^{C} &=-\frac{1}{2}[\,\bm{I}_i^{-1}(\bm{r}_i\times 
\frac{1}{w_i + w_j}C^{C}\bm{n}),0]\bm{q_{i}}.\\
\label{eqa:rp_solve}
\end{aligned}
\end{equation}
For more details, please refer to \cite{muller2020detailed}.

The following constraint we address is the shear-stretch and bend-twist constraints used for elastic rods (i.e. \ac{DLO}).
To describe these constraints, an orthogonal frame with basis $\{{\bm{d}_1(\bm{q})},\bm{d}_2(\bm{q}),\bm{d}_3(\bm{q})\}$ is attached to the center of each rod segment, where the vectors are denoted as directors.
$\bm{q}$ represents the rotation from a fixed world coordinate system with basis $\{{\bm{e}_1},\bm{e}_2,\bm{e}_3\}$ to the local frame.
The third director $\bm{d}_3(\bm{q})=\bm{R}(\bm{q})\bm{e}_3$ is parallel to the normal direction of each rod segment.
The shear-stretch constraint for two adjacent particles at positions $\bm{x}_i$ and $\bm{x}_j$ is then given by
\begin{equation}
\bm{C}^{S}(\bm{x}_i, \bm{x}_j, \bm{q}) 
= \frac{1}{l_{ij}}(\bm{x}_j - \bm{x}_i)
% - \bm{R}(\bm{q})\,\bm{e}_3 = \bm{0},
-\bm{d}_3(\bm{q}) = \bm{0},
\end{equation}
where $l_{ij}$ is the length of the segment at its rest position.
This constraint maintains the distance of each segment at its rest length and ensures that the direction of the tangent aligns with the rod segment's $\bm{d}_{3}(\bm{q})$ direction.
For faster computations, we simplify the inertia matrix for each rod segment into a single scalar $m_{q}$ following \cite{deul2016position}.
The resulting state updates are as follows:
\begin{equation}
\label{eqa:solve_shear}
\begin{aligned}
\Delta \bm{x}_i &= \frac{m_i^{-1} l_{ij}}{m_i^{-1} + m_j^{-1} + 4 m_q^{-1}l_{ij}^{2}} 
% \left( \frac{1}{l_{ij}} \left( \bm{x}_j - \bm{x}_i \right) - \bm{d}_3 \right), \\
\bm{C}^{S}(\bm{x}_i, \bm{x}_j, \bm{q}), \\
\Delta \bm{x}_j &= - \frac{m_j^{-1} l_{ij}}{m_i^{-1} + m_j^{-1} + 4 m_q^{-1}l_{ij}^{2}} 
% \left( \frac{1}{l_{ij}} \left( \bm{x}_2 - \bm{x}_1 \right) -  \bm{d}_3 \right), \\
\bm{C}^{S}(\bm{x}_i, \bm{x}_j, \bm{q}), \\
\Delta \bm{q} &= \frac{2 m_q^{-1} l_{ij}^2}{m_i^{-1} + m_j^{-1} + 4 m_q^{-1}l_{ij}^{2}} 
% \left( \frac{1}{l_{ij}} \left( \bm{x}_2 - \bm{x}_1 \right) - \bm{d}_3 \right) \bm{q} \bm{e}_3.
\bm{C}^{S}(\bm{x}_i, \bm{x}_j, \bm{q}) \bm{q} \bar{\bm{e}}_3.
\end{aligned}
\end{equation}
Here $\bar{\bm{e}}_3$ is the quaternion representation of ${\bm{e}}_3$. Finally, we have the bend-twist constraint 
\begin{equation}
\begin{aligned}
\bm{C}^{B}(\bm{q}_i, \bm{q}_j) 
&= \mathfrak{Im}( \bar{\bm{q}}_i \bm{q}_j - \bar{\bm{q}}_{i}^{0} \bm{q}_{j}^{0} \big) 
= \bm{\Omega}-\alpha\bm{\Omega}^{0}
= \bm{0} \\
\alpha &= \operatorname{sign}(\bm{\Omega}+\bm{\Omega}^{0}),
\end{aligned}
\end{equation}
where $\bm{\Omega}$ is the discrete Darboux vector $\bm{\Omega}=\frac{2}{l_{ij}} \, \mathfrak{Im} \big[ \bar{\bm{q}}_i \bm{q}_{j} \big]$, with $l_{ij}$ being the average length of two segments and $\mathfrak{Im}[\cdot]$ refers to the imaginary part of a quaternion. The bend-twist constraint 
moves the state of the rod such that its curvature and bending match its rest shape described by $\bm{\Omega}^{0}$. 
Here, the displacements of the particles are given by:
\begin{equation}
\label{eqa:solve_bend}
\begin{aligned}
\Delta \bm{q}_i &= \frac{m_{q_i}^{-1}}{m_{q_i}^{-1} + m_{q_j}^{-1}} \bm{q}_j \bm{C}^{B}(\bm{q}_i, \bm{q}_j), \\
\Delta \bm{q}_j &= - \frac{m_{q_j}^{-1}}{m_{q_i}^{-1} + m_{q_j}^{-1}} \bm{q}_i \bm{C}^{B}(\bm{q}_i, \bm{q}_j).
\end{aligned}
\end{equation}

Since the Cosserat rod model explicitly represents bending and twisting through a continuum mechanics formulation \cite{kugelstadt2016position}, it provides a meaningful physical description of rope deformation.
This yields more physically consistent deformation and contact reactions than shape matching, which enforces geometric constraints without modeling the underlying mechanics.
Furthermore, by making the parameterization of the Cosserat rod model explicit, we can enable system identification, which we plan to explore in future work.

% ===============================================================================
\subsection{Scene Initialization}
Given multiple RGB-D images from various viewpoints, we use SAM2 \cite{ravi2024sam} to extract instance masks of the object.
A bounding box is generated, and we fill it with evenly distributed spheres of 5 mm radius in 3D space. 
We also segment the workspace surface and use RANSAC \cite{fischler1981random} to fit a plane over it. This plane serves as the ground plane in the simulation.
For each sphere center representing the object, we project it onto the image planes.
We discard points that fall outside the mask, below the ground plane, or have a depth smaller than the corresponding pixel value.
The remaining spheres are then used to approximate the object.
Given the density of the rigid body, its mass, center of mass, pose, and inertial matrix are computed by integrating over the remaining spheres. 
The object's density can be obtained either by querying the VLM or by using an online system identification. 
In this work, we assume the density is known and leave automated estimation to future work.
As in \cite{abou2025real}, we also use these spheres for collision detection in the simulation.
For a DLO, we first train its 3D Gaussian representation. 
Then, we use it to render an image from a top view.
The 2D skeleton of the rope is extracted from the rendered image using Lee's algorithm \cite{lee1994building}.
Finally, we order the skeleton points by finding the longest path, yielding a sequence of evenly spaced spheres along the centerline. 
The radius of each sphere is computed from the point cloud.
The physical properties used to set up the rope model in PBD are estimated based on the material, radius, and length.

After obtaining the necessary information to set up the PBD simulation, we initialize the 3D Gaussians from the segmented object point cloud and optimize them using the photometric loss as described in \cite{kerbl20233d}. 
We train the Gaussians using the Adam optimizer for 1000 steps.
The learning rates are set $1e^{-5}$ for means, $1e^{-3}$ for quaternions, $2.5e^{-3}$ for colors, $2e^{-3}$ for scales, and $5e^{-2}$ for opacities.
Finally, the optimized Gaussians are anchored to the nearest sphere of the object.
For the robot, we manually populate each link with spheres for collision checking in PBD and store the corresponding optimized 3D Gaussians. 

\begin{figure}
\centering
\includegraphics[width=0.35\columnwidth]{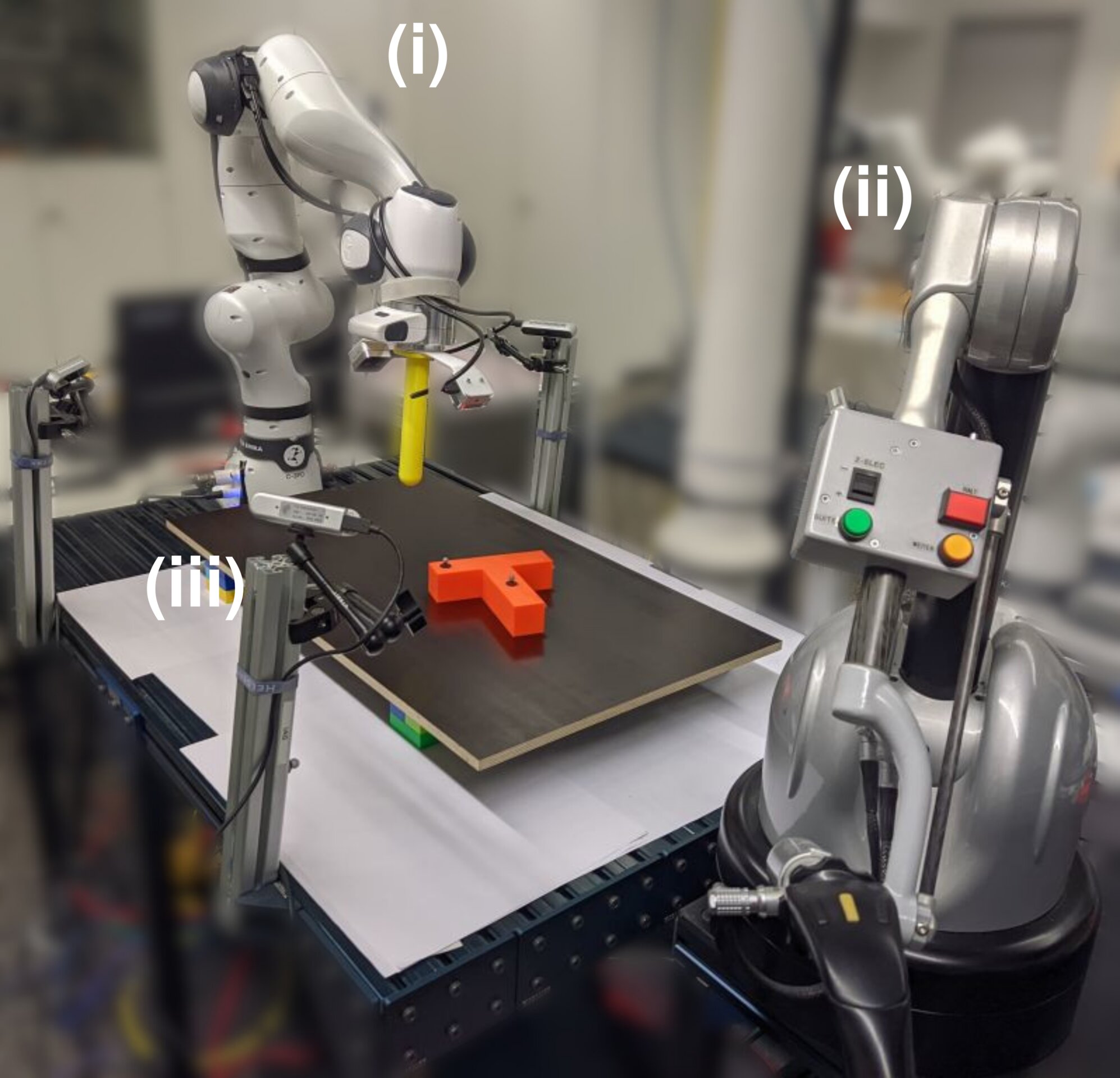}
\includegraphics[width=0.45\columnwidth]{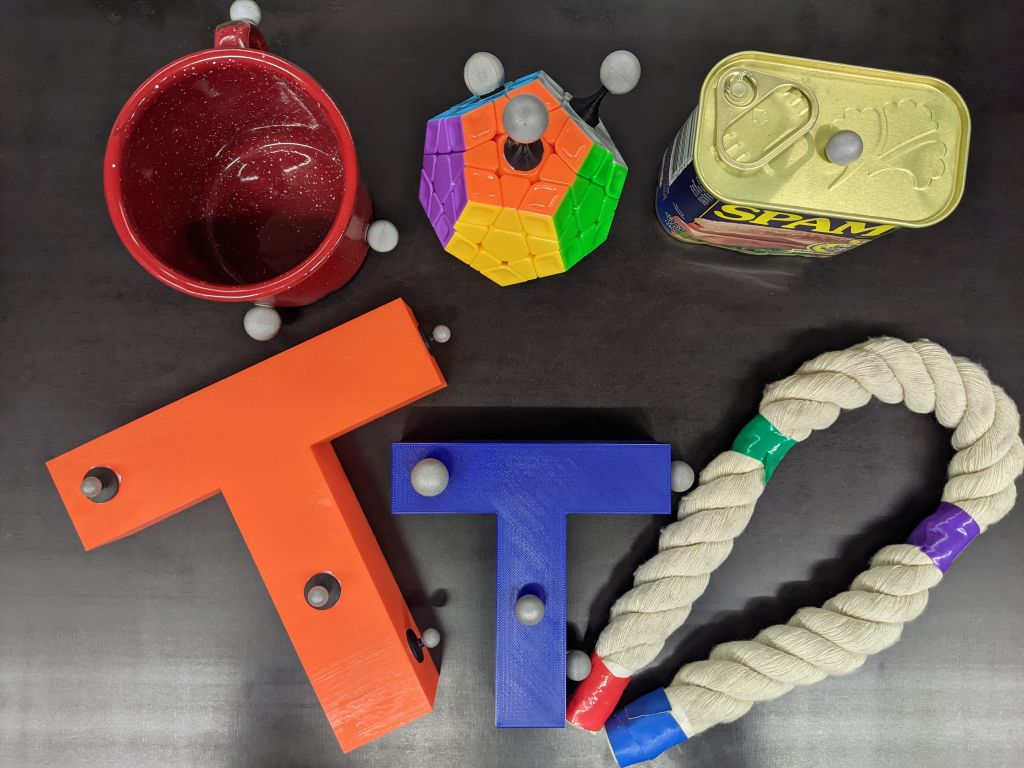}
\label{fig:stetup}
\hfill
\caption{On the left, we show the experimental setup, with the (i) Franka Research 3 robot with the custom end effector tool, (ii) the haptic interface for teleoperation, and (iii) the set of Intel RealSense D415 scene mounted cameras. On the right, we show the set of all objects used for real-robot experiments.}
\label{fig:experimental_setup}
\end{figure}

\subsection{Online Tracking}
After scene initialization, the objects with bonded Gaussians are loaded into the PBD simulation. Online tracking then proceeds in two stages: a prediction step and a correction step. The system operates at 25 Hz. 
After receiving the current RGB images and robot configurations, we initiate the prediction step by setting the robot links' poses in simulation to those calculated from the intermediate joint configuration using its forward kinematics. Next, a PBD step, as described in Section~\ref{sec:priliminary_PBD}, is executed, and the Gaussians are moved coherently with the predicted state. 
We implement the PBD using NVIDIA warp \cite{warp2022}, and each simulation step takes 0.1 ms on a single NVIDIA 4090.

Subsequently, three images are rendered according to the calibrated camera configurations using the 3D Gaussians.
Since only the Gaussians of the robot and the object are saved for rendering, we run EfficientTAM \cite{xiong2025efficient} to obtain the segmentation of the object and remove the background of the ground truth image.
Instead of optimizing the Gaussians to move separately, we force them to move rigidly. 
We apply a transformation $T \in \bm{SE}(3)$ on the means and covariance matrices of the object Gaussians.
The photometric loss is calculated by comparing the mean squared error between the rendered images and the received camera images.
The transformation is optimized to minimize photometric loss using 6 Adam optimization steps \cite{kingma2014adam}.
The correction force on each particle is calculated by $\bm{f}_i = K_p(\sum_{j}T(\bm{\mu}_j) - \bm{\mu}_j)/N_i$, where $K_p$ is a tunable gain, $T(\bm{\mu}_j)$ is the transformed Gaussians mean, and $N_i$ is the number of Gaussians bonded to each particle.
The correction force and moment for each object are computed by aggregating the particle forces as $\bm{f}=\sum_i\bm{f}_i$ and $\bm{\tau}=\sum_i\bm{f}_i\times\bm{r}_i$, where $\bm{r}_i$ is the vector from the center of mass to the position of the transformed Gaussian.
Finally, the robot links' poses are set according to the current joint configuration, the correction forces are added to the simulation, and one PBD simulation step is executed.
Here, the object segmentation takes approximately 24 ms, pose optimization requires 10 ms, and the remaining 6 ms are spent on simulation and I/O, resulting in a total latency of about 40 ms.

%% file: sections/5_experiments.tex
\section{Experimental Results}
\label{sec:experiments}
In this section, we validate our \ac{GT} framework. To analyze its performance, we design a set of experiments to answer the following question: (i) \textit{What matters the most for reliable object tracking, to have rigid body dynamics models or to perform shape matching between time steps?}; (ii) \textit{Does adding shape segmentation improve our performance?}; (iii) \textit{Can our system handle DLOs, such as a rope?}; and finally (iv) \textit{Can we leverage \ac{GT} prediction and correction steps to perform downstream tasks such as planning? }

\subsection{Experimental Setup}

\begin{figure}
    \centering
    \includegraphics[width=1\linewidth]{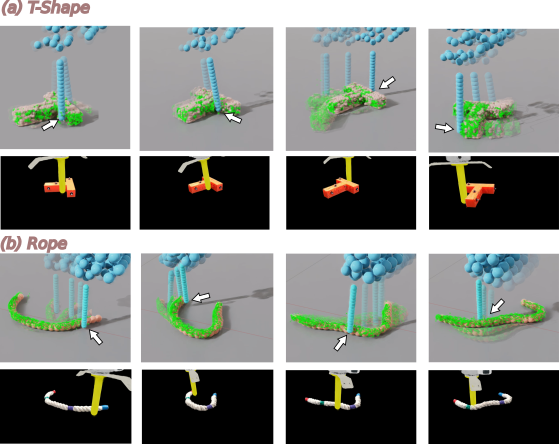}
\caption{We show qualitative results for tracking both the \rope{} and \tblock{} objects. The ground truth, overlaid on the particle simulation, is shown in green, while the simulated spheres are depicted in blue. The arrow indicates the direction of the applied force. Below the particle representation, we show rendered 3D Gaussians throughout the experiment. }
\label{fig:qual_results}
\end{figure}
\subsubsection{Datasets} 
We evaluate our methods on both the simulated dataset from~\cite{abou2024physically} and a real-world dataset. 
Here, the simulated dataset comprises three tasks: (i) single-object pushing, where a rigid object is placed on a white table and pushed horizontally by a pusher tool mounted at the robot’s end-effector; (ii) multiple-object pushing, where one T-shaped and one I-shaped block are placed on the table and pushed by the pusher; and (iii) object pushover, where a T-shaped block standing upright on the table is first pushed over and then pushed horizontally. Each task includes five scenes, with trajectories lasting 8 seconds. 
Three cameras record the scenes at 1280×720 resolution and at 25 Hz frame rate.
We note that ~\cite{abou2024physically}  also includes a deformable object pushing and the pickup task. However, as the physics prior deviated too much from the physics in the simulation, evaluation on this task was unreliable and thus not performed. 

To collect the real-world dataset, we used a seven Degrees-of-Freedom (DoF) Franka Research 3 robot equipped with a 3D-printed cylindrical pusher. Four Intel RealSense D415 cameras were mounted at the corners of the robot’s workspace to capture visual data, and a haptic interface Virtuose 6D was employed to teleoperate the robot. Here, we use all four cameras for initialization, and three for online tracking. Finally, we use OptiTrack markers for motion tracking to obtain the ground truth of the object position and orientation for evaluation. Our experimental setup is shown in Fig.~\ref{fig:experimental_setup}-left and the set of objects used in the real-world experiments in Fig.~\ref{fig:experimental_setup}-right. The objects include a \mug{}, a \spamcan{} from the YCB dataset \cite{calli2015benchmarking}, two 3D-printed \tblock{} with different sizes and colors, a \rubik{}, and a \rope{}. In the real-world dataset, four tasks similar to those in the simulation were performed. (i) single object pushing, where we teleoperated the robot to push rigid objects on the table; (ii) a pushover task, where we operated the robot to push down a standing \tblock{} to lie on the table,
(iii) rope pushing, where we teleoperated the robot to push a rope on the table,
and (iv) multiple-object pushing, where we placed two rigid objects on the table and teleoperated the robot to push them, including collisions between objects.
The length of each trajectory ranges from 30 seconds to 40 seconds.
Three cameras capture the scene at 848x480 resolution and 25 Hz framerate. Fig.~\ref{fig:qual_results} shows an example of both the \rope{} and \tblock{} pushing tasks.

\subsubsection{Baselines}
We compare our method to two different baselines. Namely,

\noindent \textbf{PEGS}: \acf{PEGS}~\cite{abou2024physically} uses shape matching in PBD and optimizes the mean and quaternion of each individual Gaussian to derive the visual correction force.

\noindent \textbf{RBD}: Follows the model from~\cite{abou2025real}. \ac{RBD} leverages rigid body physics, similarly to us. However, their Gaussians optimization procedure is done in the same manner as in \ac{PEGS}.

We note that, as our baselines do not fully work on deformable objects, we only compare on rigid-body tasks. Additionally, to verify the effectiveness of specific elements of our method, we also implemented two modified versions of our method for ablation:

\noindent \textbf{\ac{GT} (only mask)}: We provide the segmentation mask but optimize the Gaussians independently.

\noindent \textbf{\ac{GT} (only pose)}: We optimize the Gaussians coherently but do not provide the segmentation mask.

\subsubsection{Metrics}
On the simulated dataset, we evaluate the mean 3D trajectories error of known query points sampled on the object.
At each time step, the query points are transformed according to the object pose in the PBD simulation after the correction step, as the prediction.
On the real-world dataset, we compare the mean position and orientation errors of rigid bodies.
For ropes, we evaluate the IoU between the rendered images using the 3D Gaussians transformed by the state of each rope segment in the PBD simulation after the correction step, and the segmented mask of the rope from the ground truth images using SAM2 \cite{ravi2024sam}.
We report the average value over three viewpoints.

\subsection{Results and Analysis}
\begin{figure}
    \centering
    \includegraphics[width=0.8\linewidth]{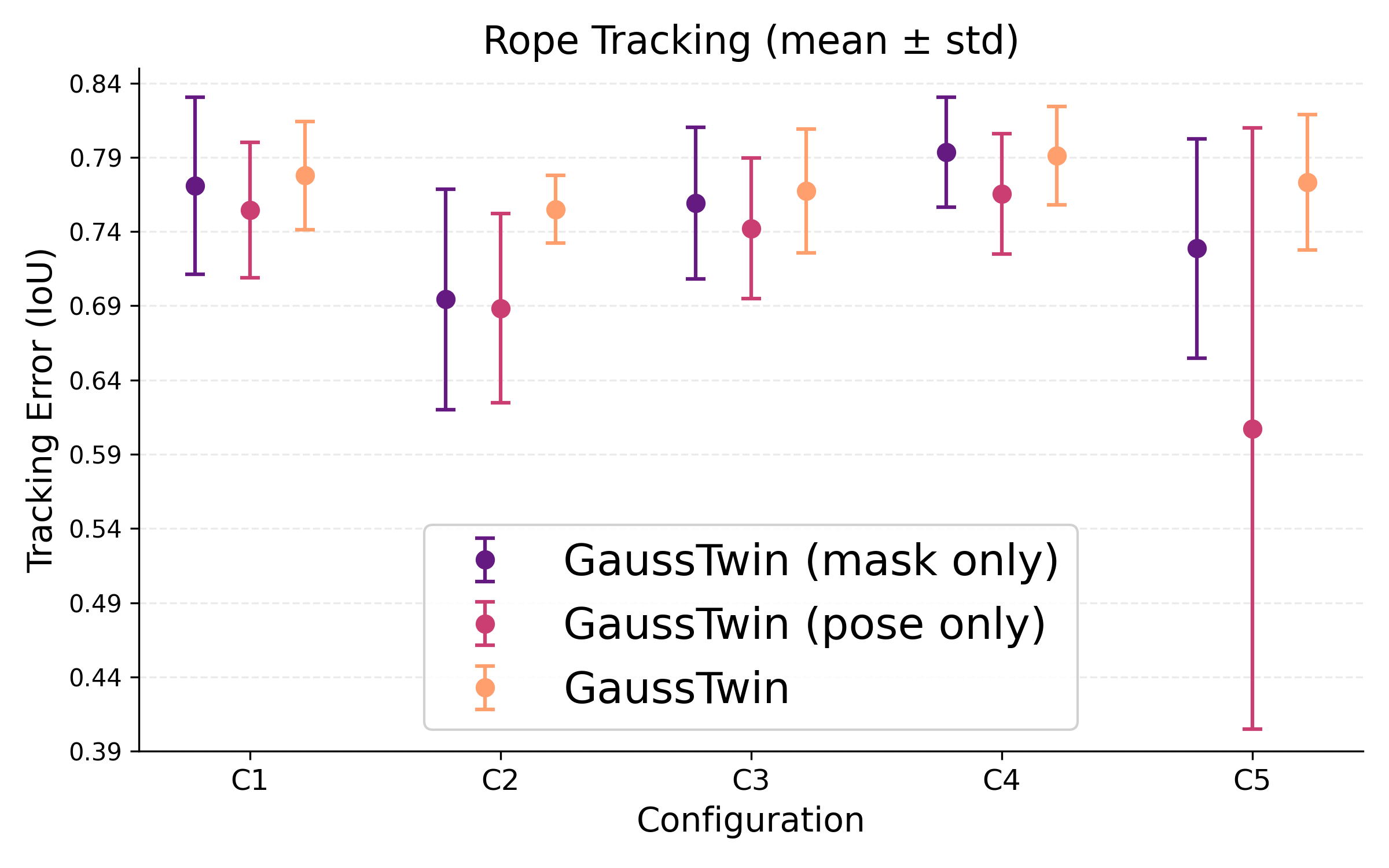}
    \caption{Error bars showing rope tracking error for five different configurations. The error is measured as the IoU between ground-truth rope pixels and the projected spheres. We compare GaussTwin with two ablations: (i) GaussTwin using only the pose, and (ii) GaussTwin using only the mask. }
    \label{fig:error_bar}
\end{figure}
\subsubsection{Simulated Dataset}
As shown in Table \ref{tab:comparison}, our method outperforms baselines in all tasks with rigid bodies. 
For all experiments, our tracking error is consistently lower and exhibits reduced variance.
Interestingly, we observe that \ac{RBD}, despite not having segmentation masks, still performs well, likely because the simulated experiments have relatively short horizons and contain fewer occlusions.
Moreover, methods that employ rigid-body dynamics generally outperform those based on shape matching, particularly in object push-down tasks, because the underlying physics simulation more faithfully captures object tumbling dynamics.
It underscores the importance of using physically plausible constraints in PBD simulation.

\input{sections/table_sim}
\subsubsection{Real-World Dataset}
\input{sections/table_real}

On the real-world dataset, we compare our method with baselines and ablations of our model.
Table \ref{tab:results} shows that our model outperforms baselines in all tasks.
Our model can stably track objects in various scenarios over a long horizon and maintain positional error within 1 cm.
Comparing tracking error with and without masks, we observe that segmentation masks improve tracking performance, particularly for orientation tracking, at the cost of 24 ms of additional segmentation time.
Nevertheless, the total latency remains 36 ms, which is suitable for real-time usage.
In the single-object pushing task, models that do not use the segmentation mask all fail to follow the object over a long horizon.
Comparing variants of method using rigid body dynamics, we find that coherently optimizing the Gaussians yields lower mean error and variance across all scenarios, thereby verifying the effectiveness of our optimization strategy.
Furthermore, most methods fail to track the orientation of objects correctly in the multiple-object pushing task because the scene includes symmetric objects, such as the mug and Rubik’s Cube. 
While best among all methods, the rotation error of GaussTwin is still relatively large, since it is difficult to correct errors using the photometric loss on textureless, symmetric objects once their orientation deviates too far from ground truth.   

For the rope-pushing scenario, our model successfully tracks the dynamic deformation of the rope under various pushing trajectories, which is shown in Fig. \ref{fig:error_bar}.
Comparing the IoU of our model and its variations, we observe that without the mask, the tracking performance drops significantly. 
Although the IoU is affected by the fact that the rendered images from 3D Gaussians occupy more pixels than the ground truth mask, we still achieve a value over 0.75 in all scenarios.
It is noteworthy that optimizing the Gaussians coherently for each segment of the rope also improves the tracking performance for deformable objects.
We also show the Gaussian rendering effect during tracking in Fig. \ref{fig:qual_results}.

\subsubsection{Model-Based Planning}
We also evaluated our particle-based dynamics model (without visual correction) for planning a sequence of pushes to align a T-shaped object with randomly sampled positions and orientations. We parameterized each push as a four-dimensional action specifying the initial and final end effector position relative to the object center, where we used polar coordinates for the initial position and planar Cartesian coordinates for the final position. We specified a reward function consisting of five terms: two for penalizing the squared position and absolute yaw error, a penalty for starting the push too close to the object, to prevent initializing the push inside of the object, a Gaussian prior on the polar-coordinate radius at $r=0.15$, and a Gaussian zero mean prior on the final position. To optimize a push, we used GMMVI~\cite{Arenz2023} to learn a 2-component Gaussian mixture model to approximate a target distribution with log-densities given by the reward function and executed a sample with high reward. We stopped the pushing sequence when no push that led to a significant improvement was found, which typically happened after two pushes. As shown in Table~\ref{tab:reward_yaw_summary}, the planned motion sequence could align the object with a position error of around 1 cm.

\input{sections/table_planning}

%% file: sections/table_sim.tex
\begin{table}
\centering
\begin{threeparttable}
\caption{Baseline comparison on the dataset from~\cite{abou2024physically}}
\label{tab:comparison}
\small 
\begin{tabular}{lccc}
\toprule
Method & Push$^\dagger$ & Push-Down $^\dagger$ & Multi-object \\
\midrule
\ac{PEGS}+M$^*$ & 0.59 $\pm$ 0.61 & 3.72 $\pm$ 2.43 & 0.58 $\pm$ 0.81 \\
\ac{PEGS}+M+P$^{*}$ & 0.54 $\pm$ 0.57 & 3.40 $\pm$ 2.12 & 0.44 $\pm$ 0.62 \\
RBD  & 0.42 $\pm$ 0.45 & 1.21 $\pm$ 0.94 & 0.48 $\pm$ 0.74 \\
GaussTwin & \reshighlight{0.34 $\pm$ 0.35} & \reshighlight{0.86 $\pm$ 0.60} & \reshighlight{0.38 $\pm$ 0.39} \\
\bottomrule
\end{tabular}
\normalsize
\begin{tablenotes}
\footnotesize
\item Each entry is reported as mean in cm $\pm$ standard deviation.
\item $^\dagger$ push and push-down tasks are reported for single objects on the scene.
\item $^*$ \ac{PEGS}+M and \ac{PEGS}+M+P follow our ablations with the additional masking and pose elements.
\end{tablenotes}
\end{threeparttable}
\end{table}

%% file: sections/table_real.tex
\begin{table*}
\centering
\small
\setlength{\tabcolsep}{3.4pt}
\renewcommand{\arraystretch}{1.05}
\begin{threeparttable}
\caption{Comparison of Methods on Different Real-World Tasks}
\label{tab:results}
\begin{tabular}{lccccccccc}
\toprule
& \multicolumn{3}{c}{Single Object Push} & \multicolumn{3}{c}{Single Object Push Down} & \multicolumn{3}{c}{Multi-object} \\
\cmidrule(lr){2-4} \cmidrule(lr){5-7} \cmidrule(lr){8-10}
Method & TE ($\pm$ std) & RE ($\pm$ std) & Lat. 
       & TE ($\pm$ std) & RE ($\pm$ std) & Lat.
       & TE ($\pm$ std) & RE ($\pm$ std) & Lat. \\
\midrule
SM~\cite{abou2024physically}
& 3.39 $\pm$ 2.18 & 33.64 $\pm$ 23.61 &  11
& 1.12 $\pm$ 0.91 &  6.19 $\pm$  5.84 &  11 
& 5.65 $\pm$ 4.77 & 29.35 $\pm$ 26.43 &  13 \\
RBD~\cite{abou2025real}
& 3.49 $\pm$ 2.84 & 17.56 $\pm$ 20.01 &  \reshighlight{10} 
& 0.99 $\pm$ 0.43 &  5.41 $\pm$  4.77 &  \reshighlight{10} 
& 4.06 $\pm$ 3.46 & 18.86 $\pm$ 18.97 &  \reshighlight{12}  \\
GaussTwin (only mask)
& 0.60 $\pm$ 0.42 &  4.85 $\pm$  3.32 &  35 
& 0.91 $\pm$ 0.45 &  4.71 $\pm$  3.50 &  35 
& 2.08 $\pm$ 1.91 & 15.32 $\pm$ 15.08 &  40 \\
GaussTwin (only pose)
& 3.40 $\pm$ 4.34 & 18.72 $\pm$ 20.31  &  11
& 1.15 $\pm$ 0.49 &  4.95 $\pm$  3.43  &  11
& 10.18 $\pm$ 9.69 & 51.63 $\pm$ 42.25 &  13 \\
GaussTwin (full)
& \reshighlight{0.43 $\pm$ 0.17} & \reshighlight{3.32 $\pm$ 1.46} & 36
& \reshighlight{0.70 $\pm$ 0.32} & \reshighlight{3.01 $\pm$ 1.80} & 36
& \reshighlight{0.85 $\pm$ 0.45} & \reshighlight{7.87 $\pm$ 6.11} & 42 \\
\bottomrule
\end{tabular}
\normalsize
\begin{tablenotes}
\footnotesize
\item TE = Mean translation error in cm across multiple objects and the entire trajectory (± standard deviation).
\item RE = Mean rotation error in degrees and the entire trajectory (± standard deviation).
\item Lat. = Mean runtime latency in ms.
\end{tablenotes}
\end{threeparttable}
\end{table*}

%% file: sections/table_planning.tex
\begin{table}[h]
\centering
\begin{threeparttable}
\caption{Mean and standard deviation of reward, position and yaw errors after the first and final push.}
\label{tab:reward_yaw_summary}
\footnotesize
\begin{tabular}{lccc}
\toprule
 & Reward & Pos. error (cm)  & Yaw error (rad.) \\
\midrule
1st Push & $-41.5 \pm 41.2$ & $1.4 \pm 0.7$ & $0.37 \pm 0.42$ \\
Last Push & $-3.48 \pm 2.92$ & $1.2 \pm 0.7$ & $0.01 \pm 0.01$ \\
\bottomrule
\end{tabular}
\normalsize
\begin{tablenotes}
\footnotesize
\item Each entry is reported as mean $\pm$ standard deviation.
\end{tablenotes}
\end{threeparttable}
\end{table}

%% file: sections/6_conclusion.tex
\section{Conclusion and Future Work}
In this work, we introduced \ac{GT}, a hybrid framework that unifies position-based dynamics with 3D Gaussian splatting to enable real-time digital twins for both rigid and deformable objects. Our evaluation was structured around four guiding questions.
First, we asked whether reliable tracking depends more on rigid-body dynamics or shape matching. Our results show that physically grounded rigid-body dynamics significantly outperform shape-matching approaches, especially for tasks involving complex object motions.
Second, we asked if adding segmentation masks improves performance. We found that segmentation consistently enhances both positional and rotational accuracy, confirming the value of visual feedback in maintaining stability over long horizons.
Third, we examined whether GaussTwin can handle deformable linear objects such as ropes. Our experiments demonstrate that it can robustly track and correct DLOs, validating the effectiveness of extending PBD with Cosserat rod models and coherent Gaussian optimization.
Finally, we asked whether GaussTwin’s prediction–correction loop can support downstream tasks such as planning. We showed that it enables model-based push planning with centimeter-level accuracy, underscoring its utility for real-world manipulation.
By explicitly addressing these four questions, we establish GaussTwin as a unified and physically consistent digital twin framework that bridges the real-to-sim gap. 
For future work, we plan to extend GaussTwin with automatic parameter estimation and integrate it into vision-based policy learning pipelines.